\documentclass{article}

\usepackage{geometry}
\usepackage{setspace}
\usepackage[numbers]{natbib}

\usepackage[separate-uncertainty = true,multi-part-units = repeat]{siunitx}
\usepackage[utf8]{inputenc} 
\usepackage[T1]{fontenc}    
\usepackage{hyperref}       
\usepackage{url}            
\usepackage{cancel}
\usepackage{booktabs}       
\usepackage{amsfonts}       
\usepackage{nicefrac}       
\usepackage{microtype}      
\usepackage{xcolor}         
\usepackage{natbib}
\usepackage{amsmath}
\usepackage{wrapfig}
\usepackage[pdftex]{graphicx}

\usepackage{svg}
\title{ Learning Nonparametric Volterra Kernels with Gaussian Processes}
\author{%
   Magnus Ross,\hspace{1mm} Michael T. Smith,\hspace{1mm} Mauricio A. {\'Alvarez} \\
    Department of Computer Science\\
  University of Sheffield\\
  Sheffield, UK, S1 4DP \\
  \texttt{\{mross1, m.t.smith, mauricio.alvarez\}@sheffield.ac.uk}\\
}

\begin{document}

\maketitle

\begin{abstract}
  This paper introduces a method for the nonparametric Bayesian
  learning of nonlinear operators, through the use of the Volterra
  series with kernels represented using Gaussian processes (GPs),
  which we term the nonparametric Volterra kernels model (NVKM). When
  the input function to the operator is unobserved and has a GP prior,
  the NVKM constitutes a powerful method for both single and multiple
  output regression, and can be viewed as a nonlinear and
  nonparametric latent force model. When the input function is
  observed, the NVKM can be used to perform Bayesian system
  identification. We use recent advances in efficient sampling of
  explicit functions from GPs to map process realisations through the
  Volterra series without resorting to numerical integration, allowing
  scalability through doubly stochastic variational inference, and
  avoiding the need for Gaussian approximations of the output
  processes. We demonstrate the performance of the model for both
  multiple output regression and system identification using standard
  benchmarks.
\end{abstract}

\section{Introduction}

Gaussian processes (GPs) constitute a general method for placing
prior distributions over functions, with the properties of samples
from the distribution being controlled primarily by the form of the covariance
function \citep{gpmlbook}. Process convolutions (PCs) are one powerful
method for building such covariance functions \citep{Barry:blackbox96,
  higdon2002space,alvarez2012kernels}. In the PC framework, the function we wish to
model is assumed to be generated by the application of some
convolution operator to a base GP with some simple covariance, and
since linear operators applied to GPs result in GPs, the result is
another GP with a covariance we deem desirable. PCs allow models for
multiple correlated output functions to be built with ease, by
assuming each output is generated by a different operator applied to
the same base function, or set of functions \citep{verHoef:convolution98, higdon2002space}.

The PC framework unifies a number of different ideas in the GP
literature. Latent force models (LFMs) \citep{alvarez2009latent} use
PCs to include physics based inductive biases in multiple output GP
(MOGP) models by using the Green's function of a linear differential
operator as the kernel of the convolution. This leads to the
interpretation of each output as having been generated by inputting a
random latent force into a linear system, with physical properties
described by the differential operator. The Gaussian process
convolution model (GPCM) of \citet{tobar2015learning} treats the
convolution kernel itself as an unknown function to be inferred from
data, and places a GP prior over it. Linear systems are entirely
described by their Green's function, so one can interpret the GPCM as
a nonparametric, linear LFM, in which the form of the system itself is
inferred from data.

In the physical world, nonlinear systems are the norm, and linearity
is an approximation. As a consequence it is desirable when dealing
with physical data to have models that can incorporate nonlinearity
naturally. Often however, given a certain set of data, it is not clear
exactly what form this nonlinearity takes, and so introducing specific
parametric non-linear operators can be overly
restrictive. \citet{alvarez2019non} present a model known as the
nonlinear convolved MOGP (NCMOGP) which introduces nonlinearity to
MOGPs via the Volterra series \citep{cheng2017volterra}, a nonlinear
series expansion used widely for systems identification, whose
properties are controlled by a set of square integrable functions of
increasing dimensionality known as the Volterra kernels (VKs). The
NCMOGP assumes these functions are both separable and homogeneous, and
of parametric Gaussian form, it also approximates the outputs as GPs
in order to make inference tractable. 

The present work introduces a
new model which drops the separability and homogeneity assumptions on
the VKs, allows their form to be learned directly from data, and makes
no approximation on the distribution of the outputs. We refer to it as
the nonparametric Volterra kernels model (NVKM).
We develop a fast sampling method for the NVKM which
leverages the recent results of \citet{wilson2020efficiently} on the
sampling of explicit functions from GPs to analytically map function
realisations through the Volterra series, avoiding the need for
computationally expensive and inaccurate high dimensional numerical
integration. Fast sampling allows for the application of doubly
stochastic variational inference (DSVI) \citep{titsias2014doubly}
for scalable learning.

The NVKM is well suited to both single and multiple output regression
problems, and can be thought of as an extension of the GPCM to both
non-linear systems and multiple outputs. The NVKM can also be
interpreted as a non-linear LFM in which the operator
is learned directly from data. We additionally present a variation of the NVKM that can be used for Bayesian systems identification, where the task is to learn operator mappings between observed input and output data, and show that it allows for considerably better quantification of uncertainty than competing methods which use recurrence \citep{mattos2017recurrent}.
\section{Background}

In this section we give a brief introduction to the mathematical background of PCs and the Volterra series.
\paragraph{Process convolutions}\label{processconvs}
In the PC framework, the set of output functions $\{f_d(t)\}^{D}_{d=1}$, with $D$ being the number of outputs, is generated by the application of some set of linear operators, specifically convolution operators, to a latent function $u$ represented by a Gaussian process, $f_d(t) = \int_{\mathcal{T}} G_d(t-\tau)u(\tau)d\tau$, where $\mathcal{T}$ is the domain of integration, and the function $G_d$ is known variously as the convolutional kernel, smoothing kernel, impulse response or Green's function, depending on context. This function must be square integrable to ensure the output is finite. A linear operator acting on a GP produces another GP \citep{gpmlbook}, and so we obtain $D$ distinct GPs. Since the latent function $u$ is shared across the outputs, these $D$ GPs are correlated, allowing joint variations to be captured, whilst the convolution with $G_d$ adapts $u$ to each output. \citet{alvarez2012kernels} show that many MOGP models can be recast in terms of the PC framework by particular choices of $G_d$ and $u$.  In LFMs, $G_d$ is taken to be the Green's function of some differential operator. We can then interpret each output as resulting from a shared random force being fed into a distinct linear system, represented by the differential operator. The smoothing kernels are usually taken to have parametric, often Gaussian form. \citet{tobar2015learning} use the PC framework for a single output, and make the smoothing kernel itself a GP, giving rise to the GPCM. \citet{bruinsma2016generalised} extends the GPCM to the multiple output case, although the model was not applied to data.
\paragraph{Voterra Series} The PC framework can be extended to represent a broader class of output functions by instead considering the outputs $\{f_d(t)\}^{D}_{d=1}$ as being the result of some non-linear system, acting on the latent function $u$. The Volterra series is a series approximation for non-linear systems that is widely used in the field of system identification \citep{cheng2017volterra}. It is given by,
\begin{equation}\label{vseries}
    f(t) = \sum_{c=1}^{C} \int_{\mathcal{T}} G_c(t-\tau_1, \ldots, t-\tau_c) \prod_{j=1}^{c} u(\tau_{j}) \text{d} \tau_{j},
\end{equation}
where $f$ is the system output, $G_c$ is the $c^{\text{th}}$ order VK,
$u$ is the system input, and $C$ is the order of the approximation. We
can think of the Volterra series as an extension of the well known Taylor expansion, that allows $f$ to have memory of the past values of $u$, that is to say $f$ depends on $u$ at all values $t \in \mathcal{T}$, and can approximate a broad class of non-linear operators. \citet{alvarez2019non} use Equation \eqref{vseries} to construct the NCMOGP, applying the Volterra series with $d$ distinct sets of VKs $\{G_{d,c}\}^{C}_{c=1}$ to a shared latent GP input $u$, to produce outputs $\{f_d(t)\}^{D}_{d=1}$. Since the Volterra series is a nonlinear operator, the output becomes an intractable, non-Gaussian process. The authors perform inference by approximating the outputs as GPs and using the first and second moments of the output process to form its mean and covariance function. To enable to computation of these moments, the authors restrict the set of VKs to those which are both separable and homogeneous, i.e. $G_{d,c}(t_1, \ldots, t_c) = \prod^c_{i=1}G_d(t_i)$. Additionally, since the moment computation requires analytically solving a number of non-trivial convolution integrals, the authors only consider a Gaussian form for the VKs.

\section{The nonparametric Volterra kernels model}
\begin{figure}
  \centering
    \includegraphics[width=1\textwidth]{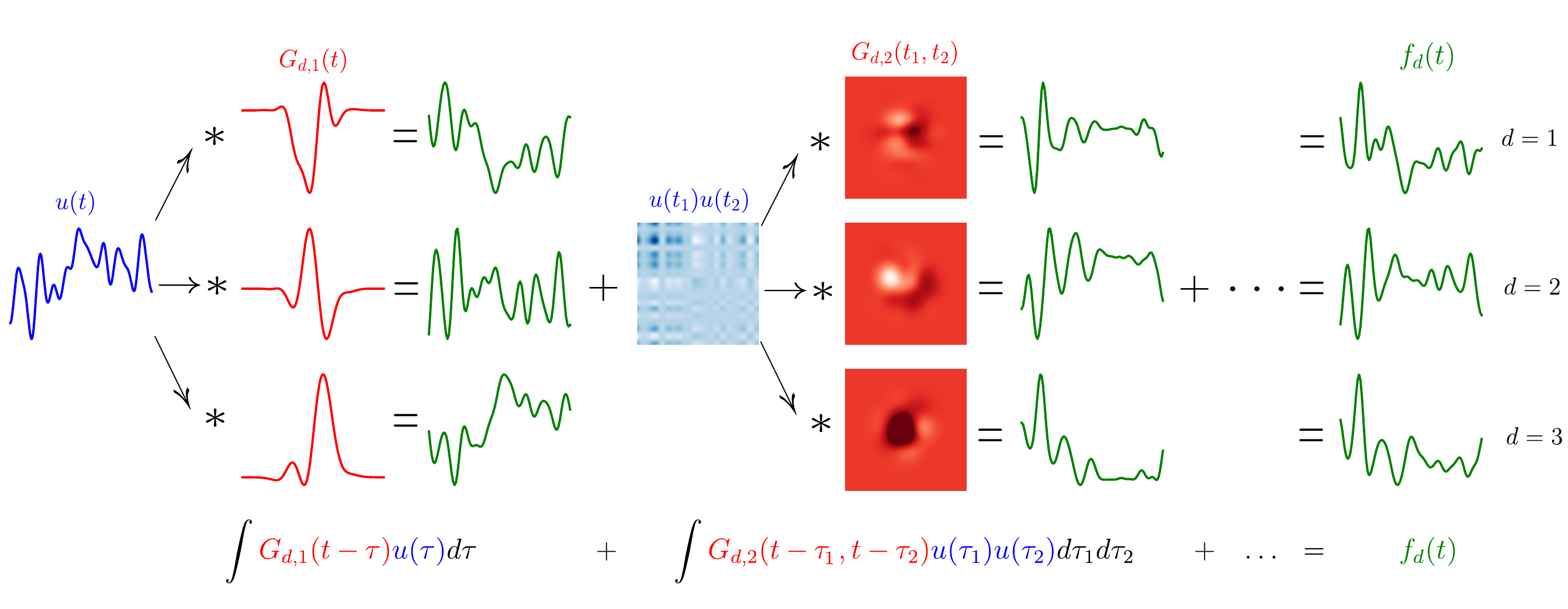}
  \caption{A diagram of the generative process for the NVKM with $C=2$
    and $D=3$, showing the stages of computation of the first order
    (on the left side) and second order (in the center) terms
    of the Volterra series for three outputs, shown in the rows, with the star representing a convolution. The 1D contribution from the second order term is obtained by taking the diagonal of the result of the 2D convolution.}\label{samplediag}
\end{figure}

The NVKM relaxes the restrictions of separability and homogeneity which are placed on the VKs in the NCMOGP, and represents these kernels as independent GPs, allowing their form and uncertainty to be inferred directly from data. The generative process for the NVKM can be stated as
\begin{equation}\label{nvkm}
    \begin{split}
        u(t) &\sim \mathcal{GP}(0, k^{(u)}(t,t')),  \\ 
        G_{d,c}(\mathbf{t}) &\sim \mathcal{GP}(0, k^{(G_{d,c})}(\mathbf{t}, \mathbf{t}')), \quad \mathbf{t} \in \mathbb{R}^c , \quad \forall c \in {1, \dots, C}, \quad \forall d \in {1, \dots, D}, \\
       f_d(t) &= \sum_{c=1}^{C} \int^\infty_{-\infty} G_{d, c}(t-\tau_1, \ldots, t-\tau_c) \prod_{j=1}^{c} u(\tau_{j}) \text{d} \tau_{j}, \\
     \end{split}
\end{equation}
where $k^{(u)}(t,t')$ is the covariance function for the input
process, and $k^{(G_{d,c})}(\mathbf{t}, \mathbf{t}')$ is the
covariance function for the $c^{\text{th}}$ VK of the $d^{\text{th}}$
output. We follow \citet{tobar2015learning} in using the decaying
square exponential (DSE) covariance for the VKs, which is
a modification to the ubiquitous square exponential (SE) covariance
that ensures the samples are square integrable. The DSE covariance has the form
\begin{equation}
k_{DSE}(\mathbf{t}, \mathbf{t}') = \sigma^2 \exp(-\alpha(\|\mathbf{t}\|^2+\|\mathbf{t}'\|^2)-\gamma\|\mathbf{t}-\mathbf{t'}\|^2),
\end{equation}
where $\|\cdot\|$ is the or $\ell^2$ norm, $\sigma$ is the amplitude, $\alpha$ controls the rate at which the samples decay away from the origin, and $\gamma$ is related to the length scale $l$ of the samples by $\gamma = \frac{1}{l^2}$. A diagram of the generative process for the model is shown in Figure \ref{samplediag}. Obtaining an exact distribution over the outputs $f_d$ is intractable, since it involves integration over nonlinear combinations of infinite dimensional stochastic processes. In order to sample from the model, and perform inference, approximations must be introduced. In particular, we employ the results of \citet{wilson2020efficiently} to sample in linear time, which enables efficient learning through the use of variational inducing points \citep{titsias2009variational} with doubly stochastic variational inference (DSVI) \citep{titsias2014doubly}.
\subsection{Sampling}
One could sample from the model by drawing from the input and filter
GPs at some finite set of locations, and then using the samples with
some method for numerical integration to find the output. As the
dimensionality of the filters increases, however, many points would be
needed to obtain an accurate answer, this quickly becomes
computationally intractable, since sampling exactly
from a GP has cubic time complexity with respect to the number of
points requested. We can sidestep this problem, and avoid the need for
any numerical integration, by representing samples from the GPs
explicitly as functions. Using the results from
\citet{wilson2020efficiently}, and following their notation, we can
write a sample from a GP $f: \mathbb{R}^c\xrightarrow[]{}\mathbb{R}$, with covariance function $k(\mathbf{t}, \mathbf{t}')$, given $M$ inducing variables $\mathbf{v} \in \mathbb{R}^M$ with
corresponding inducing inputs $\{\mathbf{z}_j\}_{j=1}^M$, with $ \mathbf{z}_j\in \mathbb{R}^{c}$, as
\begin{equation}
(f|\mathbf{v})(\mathbf{t}) = \sum^{N_b}_{i=1} w_i\phi_i(\mathbf{t}) + \sum^M_{j=1}q_jk(\mathbf{t}, \mathbf{z}_j),
\label{pathbasis}
\end{equation}
where $\{\phi_i\}^{N_b}_{i=1}$ is a random Fourier basis, with $N_b$
being the number of basis functions in the approximation,
$\mathbf{w}\in \mathbb{R}^{N_b}$ with entries $w_i \sim \mathcal{N}(0,
1)$, and $q_j$ are the entries of the vector
$\mathbf{q}=\mathbf{K}^{-1}(\mathbf{v}-\Phi\mathbf{w})$, where
$\mathbf{K}\in \mathbb{R}^{M \times M}$, with elements $K_{m,n} =
k(\mathbf{z}_n, \mathbf{z}_m)$, is the covariance matrix of the
inducing points, and $\Phi \in \mathbb{R}^{M \times N_b}$ is a feature
matrix, with each basis function being evaluated at each inducing
location.  The random Fourier basis is obtained by first sampling
$\beta_i \sim U(0, 2\pi)$, where $U$ is the uniform distribution, and then sampling  $\boldsymbol{\theta}_i \sim FT(k)$, where $FT$ is the Fourier transform of the covariance function, which is the spectral density of the process. The basis functions are then given by $  \phi_i (\mathbf{t})= \sqrt{2/N_b}\cos(\boldsymbol{\theta}^{\top}_i \mathbf{t} + \beta_i )$. We can see Equation \eqref{pathbasis} as consisting of an approximate GP prior, using a random Fourier features approximation \citep{rahami2007random}, with a correction term which uses Matheron's update rule to account for the inducing points. By using Equation \eqref{pathbasis}, samples can be obtained in linear time with respect to the number of requested points. It should be noted that Equation \eqref{pathbasis} only applies to GPs with stationary kernels, however, the DSE covariance required for $G_{c, d}$, is non-stationary. It can be shown that the process $\exp(-\alpha||\mathbf{t}||^2)G'_{c, d}(\textbf{t})$ has the DSE covariance if $G'_{c, d}$ has the SE covariance.  We can then write a sample from the output of the NVKM as,
\begin{equation}
    (f_d|\{ \mathbf{v}^G_{d,c}\}^{C}_{c=1}, \mathbf{v}^u)(t) = \sum_{c=1}^{C} \int^\infty_{-\infty}e^{-\alpha \sum^c_{i=1} (t - \tau_i)^2}(G'_{d, c}|\mathbf{v}^G_{d,c})(t-\tau_1, \ldots, t-\tau_c) \prod_{j=1}^{c} (u|\mathbf{v}^{u})(\tau_{j}) \text{d}\tau_j,
    \label{nvkms}
\end{equation}
which can be computed analytically, assuming that $u$ also has the SE
covariance, by representing the Fourier basis in complex form, and
factorising the integrals, leading to combinations of sums and
products of single dimensional integrals of the form
$\int^\infty_{-\infty} \exp (-ax^2 +bx) dx =
\sqrt{\pi/a}\exp(b^2/4a)$. See Appendix \ref{explder} for details of the computation.

\subsection{Inference}
\begin{wrapfigure}{r}{0.40\textwidth}
\vspace{5mm}
    \includegraphics[width=0.40\textwidth]{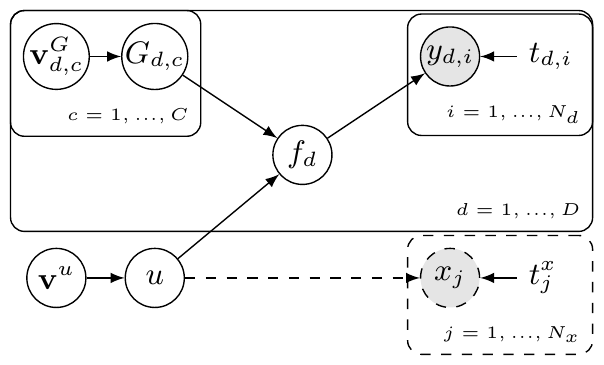}
  \caption{A graphical model for the NVKM, where the dashed elements are added to form the IO-NVKM. Note that nodes $u$, $f_d$, and $G_{d, c}$ are not random variables, they are random processes, but the distinction is not made in the diagram for the sake of clarity.}\label{gmodel_figure}
  \vspace{-4mm}
\end{wrapfigure}

Learning with the NVKM implies making inference of the input
  process $u$, along with all VKs $\{G_{d,c}\}^{C, D}_{c, d=1}$, from
  observed output data $\{\mathbf{y}_d\}^{D}_{d=1}$ with $\mathbf{y}_d
  \in \mathbb{R}^{N_{d}}$, which are the functions
  $\{f_d\}^{D}_{d=1}$ evaluated at points $\{\mathbf{t}_d\}^{D}_{d=1}$
  with $\mathbf{t}_d \in \mathbb{R}^{N_{d}}$, corrupted by some i.i.d
  Gaussian noise. That is to say $y_{d, i} = f_d(t_{d,i}) +
  \epsilon_{d,i}$ with $\epsilon \sim \mathcal{N}(0,
  \sigma^2_{y_d})$. Let $\mathbf{v}^G_{d,c}=G_{d,c}(\mathbf{z}^G_{d,c})$ denote the inducing points for the VKs, and $\mathbf{v}^u=u(\mathbf{z}^u)$ denote the inducing points for the input. The joint distribution over these inducing points and the latent functions then has the following form
\begin{equation}\label{bigprob}
    p(\{\mathbf{y}_d\}^{D}_{d=1},\{G_{d,c}, \mathbf{v}^G_{d,c}\}^{C, D}_{c, d=1} , u, \mathbf{v}^u) =\prod^{D, N_d}_{d,i=1}  p(y_{d,i}|f_d(t_{d,i}) )
\prod^{D,C}_{d,c=1} p(G_{d,c} | \mathbf{v}^G_{d,c})p(\mathbf{v}^G_{d,c})p(u | \mathbf{v}^u) p(\mathbf{v}^u),
\end{equation}
where $f_d(t_{d,i})$ depends on the VKs and input through Equation \eqref{nvkm},  the likelihood is $p(y_{d,i}|f_d(t_{d,i})) =\mathcal{N}(y_{d,i};f_d(t_{d,i}), \sigma_{y_d}^2)$ ,  $p(G_{d,c} | \mathbf{v}^G_{d,c})$ and $p(u | \mathbf{v}^u)$ are GP posterior distributions, and $p(\mathbf{v}^u)$  and $p(\mathbf{v}^G_{d,c})$ are the prior distributions over the inducing points. The dependency structure of the model is described in Figure \ref{gmodel_figure}. We form an approximate variational distribution, in a similar way to \citet{tobar2015learning}, using a structured mean field approximation. That is to say, we mirror the form of the true joint distribution, and replace the prior distributions over the inducing points with variational distributions, $q(\mathbf{v}^G_{d,c})$ and $q(\mathbf{v}^u)$, leading to the variational distribution
\begin{equation}
q(\{G_{d,c}, \mathbf{v}^G_{d,c}\}^{C, D}_{c, d=1} , u, \mathbf{v}^u) =  \prod^{D,C}_{d,c=1} p(G_{d,c} | \mathbf{v}^G_{d,c})q(\mathbf{v}^G_{d,c})p(u | \mathbf{v}^u)q(\mathbf{v}^u).
\end{equation}
The optimal form of the variational posteriors are multivariate Gaussians, $q(\mathbf{v}^{u}) = \mathcal{N}(\mathbf{v}^{u};\boldsymbol{\mu}^{u}, \boldsymbol{\Sigma}^{u})$  and $q(\mathbf{v}^G_{d,c}) = \mathcal{N}( \mathbf{v}_{d,c}^G; \boldsymbol{\mu}^G_{d,c}, \boldsymbol{\Sigma}^G_{d,c})$, where the mean vectors and covariance matrices of these distributions are variational parameters. This form of variational approximation leads to a variational lower bound,
\begin{equation}\label{bound}
    \mathcal{F} =\sum^{D, N_d}_{d,i=1} \mathbb{E}_q[\log p(y_{d,i}|f_d(t_{d,i}))] - \sum^{D, C}_{d,c=1} \text{KL}[q(\mathbf{v}^G_{d,c}))||p(\mathbf{v}^G_{d,c}))] -\text{KL}[q(\mathbf{v}^u)||p(\mathbf{v}^u)],
\end{equation}
where $\text{KL}[.||.]$ represents the Kullback-Lieber (KL)
divergence. The expression above is optimised using gradient
descent. The KL divergences have closed form. The derivation of
the bound and KL divergences are given in Appendix \ref{boundder}. The expectation of the log likelihood of the outputs, given in the first term, is intractable, due to the nature of nonlinearity introduced by the Volterra series. We instead compute a stochastic estimate of the log likelihood by sampling from the model and using,
\begin{equation}\label{loglikeapprox}
    \mathbb{E}_q[\log p(y_{d,i}|f_d(t_{d,i}))] \approx \frac{1}{S} \sum^S_{s=1} \log p(y_{d,i}|(f_d|\mathbf{v}^G_{d,c}, \mathbf{v}^u)(t_{d,i})),
\end{equation}
where $\mathbf{v}^G_{d,c}$ and $\mathbf{v}^u$ are first sampled from
their respective variational distributions, and then used in
Equation \eqref{nvkms} to generate a
sample from $f_d$. To make the inference scheme scalable, we compute
the bound on randomly sub-sampled mini batches of the data set, which
alone is known as stochastic variational inference
\citep{hoffman2013stochastic}. When this source of stochasticity is
combined with the stochastic estimate of the expected log likelihood,
we have DSVI \citep{titsias2014doubly}.
\par In the standard NVKM model, the input process $u$ is a latent
function with no observed data associated with it. There are many
situations in which, instead of learning a distribution over some
output functions alone, we wish to learn an operator mapping between an input function and an output function, or functions. That is to say, in addition to observing data $\{\mathbf{y}_d\}^{D}_{d=1}$ we also observed the input process $u$ at locations $\mathbf{t}^{x} \in \mathbb{R}^{N_x}$ corrupted with i.i.d noise, which we denote $\mathbf{x}$, so $x_i = u(t^x_i) + \epsilon^x_i$ with $\epsilon^x \sim \mathcal{N}(0, \sigma^2_{x})$ . We apply a simple modification to the inference scheme of the NVKM, in order to form a new model which we term the input/output NVKM (IO-NVKM). For the IO-NVKM, we pick up an additional likelihood term in Equation \eqref{bigprob}, with the variational distribution remaining unchanged, and we obtain a new bound,
\begin{equation}
    \mathcal{F}_{IO} = \mathcal{F} + \sum^{N_x}_{j=1} \mathbb{E}_q[p(x_{j}|u(t^{x}_{j}))],
\end{equation}
where $p(x_{j}|u(t^{x}_{j}))=\mathcal{N}(x_{j};u(t^{x}_{j}),
\sigma_x^2)$, $\mathcal{F}$ is given by Equation \eqref{bound} and
we approximate the expectation as in Equation \eqref{loglikeapprox}. The relationship between the NVKM and IO-NVKM is illustrated in Figure \ref{gmodel_figure} where the dashed sections are added to form the IO model.

\section{Related Work}

In the this section, we give a brief overview of existing ideas in the literature that have some connection to the NVKM.
\paragraph{Non-parametric kernels} 
Covariance function design and selection is key to achieving good performance with GPs. Much work
has been done on automatically determining covariance functions from
data, for example by building complex covariances by composing simpler
ones together \citep{duvenaud2011additive}, or by using deep neural
networks to warp the input space in complex ways before the
application of a simpler kernel \citep{wilson2016deep}. Flexible
parametric covariances can also be designed in frequency space
\citep{wilson2013gaussian}. Alternatively, some
efforts have been made to learn covariance functions using GPs
themselves. In addition the the GPCM, another model that learns
covariances nonparametrically is due to \citet{benton2019function},
who use GPs to represent the log power spectral density, and then apply
Bochner's theorem to convert this to a representation of the
covariance function. GPs are fully specified by their first two
moments, so by learning the covariance function and mean function, one
knows all there is to know about the process. The present work uses
the formalism of the Volterra series to learn the properties of more
complex, non-Gaussian processes, non-parametrically. We can think of
this as implicitly learning not just the first and second order
moments of the process, but also the higher moments, depending on the
value of $C$. In the case of $C=1$ and $D=1$, the NVKM and the GPCM are
the same, except for the fact the GPCM uses the white noise as the
input process, whereas we use an SE GP.

\paragraph{LFMs and MOGPs} As discussed in Section \ref{processconvs},
we can interpret the first order filter function as the Green's
function of some linear operator or system, and so by placing a GP
prior over it, we implicitly place a prior over some set of linear
systems \citep{tobar2015learning}. Since standard LFMs use an operator
of fixed form, we can interpret the NVKM in the case $C=1$, $D\geq 1$
as being an LFM in which the generating differential equation itself
is learned from data. LFMs can be extended to cases in which the
differential operator is nonlinear. \citet{hartikainen2012state}
recast a specific non-linear LFM in terms of a state space formalism
allowing for inference in linear time. \citet{NIPS2008_5487315b} use
Markov chain Monte Carlo to infer the parameters of a specific
nonlinear ODE describing gene regulation, using a GP as a latent input
function. \citet{ward2020black} use black box VI with inverse
auto-regressive flows to infer parameters of the same ODE.  When $C>1$,
the NVKM can be interpreted as a nonlinear nonparametric
LFM. \citet{alvarez2019non} use the fixed, parametric VKs to build an
MOGP model. In contrast to the NVKM, they approximate the outputs as
GPs and use analytical expressions for the moments to perform exact
GP inference.

\paragraph{Nonlinear system identification} The IO-NVKM falls into the class of
models which aim to perform system identification. A key concern of systems identification is determining how the output
of certain systems, often represented by differential operators,
respond to a given input. GPs have long been used for the
identification of both linear and nonlinear systems. Many models exist
which use GPs to recurrently map between previous and future states,
including GP-NARX models \citep{kocijan2005dynamic}, various state
space models \citep{svensson16compuatoinally,svensson2017flexible} and
recurrent GPs (RGPs) \citep{Mattos-recurrent16}. The thesis of
\citet{mattos2017recurrent} gives a summary of these
methods. \citet{WORDEN2018194} detail a method for the combination of
the GP-NARX model with a mechanistic model based on the physical
properties of the system under study, leading to improved performance
over purely data driven approaches.  The IO-NVKM differs from these
models in that instead of learning a mapping from the state of the
system at a given point to the next state, we use GPs to learn
an operator that maps the whole input function to the whole output
function.

\section{Experiments}\label{experiements}
   \begin{figure}
    \includegraphics[width=1.0\textwidth]{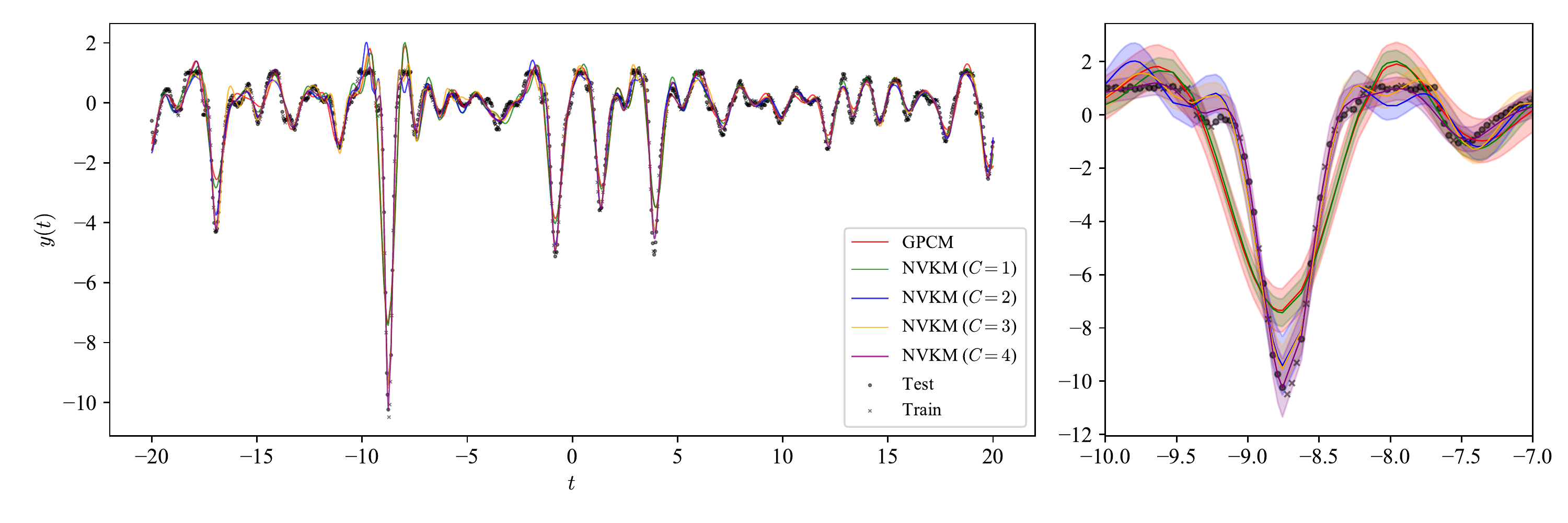}
  \caption{Model predictions on the synthetic data, with crosses indicating training points and dots indicating test points.  The right plot shows an enhanced view of the peak around $t=-9$, and the shaded regions show $2\sigma$ confidence.}
\label{toy_figure}
  \end{figure}
  
 \begin{wraptable}{r}{9cm}
 \vspace{-5mm}
  \caption{Performance on the synthetic data set, showing mean and standard deviation for 10 repeats.}
  \label{toy_table}
  \centering
  \begin{tabular}{
    lllll
  S[table-format=8.0(7)]
  }
    \toprule
    Model     & NMSE     & NLPD\\
    \midrule
GPCM          & 0.199  \textpm   0.023 & 1.080  \textpm   0.130\\
NVKM ($C=1$)  & 0.196  \textpm   0.047 & 2.084  \textpm   0.398\\
NVKM ($C=2$)  & 0.108  \textpm   0.065 & 0.638  \textpm   0.580\\
NVKM ($C=3$)  & \textbf{0.055  \textpm  0.016} & \textbf{0.124  \textpm   0.107}\\
NVKM ($C=4$)  & 0.084  \textpm   0.087 & 0.149  \textpm   0.331\\
    \bottomrule
  \end{tabular}
\end{wraptable} 

For all the following experiments we place the inducing locations for
both the input process and VKs on a fixed grid. The GP representing
the $c$th VK has input dimension $c$, which means that the number of
inducing points required to fully characterise it scales exponentially
with $c$ if they are placed on a grid. For all experiments we use 15,
10, 6 and 4 inducing points per axis for each of the 1st to 4th order
filters respectively, centered on zero. We treat the range of the
points of each VK as a hyperparameter, and fix $\alpha$ such that the
decaying part of the DSE covariance causes samples to be near zero at
the edge of the range. For $u$ we use approximately 1/10 of the number
of inducing points as data points (average number of data per output
for multi-output problems). The VK GP length scales, VK GP amplitudes,
and input GP amplitude are optimised along with the variational
parameters by maximising the variational bound using gradient
descent. For computational reasons, the input process length scale is
fixed based on the spacing of the input inducing points, and the noise
hyperparameters are fixed to a small value whilst optimisation is
taking place, and then fit alone afterwards. The model is implemented
using the Jax framework \citep{jax2018github}.\footnote{Code available at \hyperlink{https://github.com/magnusross/nvkm}{\texttt{github.com/magnusross/nvkm}}} For all experiments we
use Adam \citep{DBLP:journals/corr/KingmaB14}. All models were trained
on a single Nvidia K80 GPU.

 \subsection{Synthetic data}\label{toy_sec}
 To illustrate the advantage of including non-linearity in the model we generate a synthetic single output regression problem which includes both hard and soft nonlinearities by sampling $g$ from an SE GP with length scale 2, computing $f_i(t)=\int e^{-2\tau^2}h_i(\tau)g(t-\tau)d\tau$ for 
 $h_1(t) =\sin(6t)$, $h_2(t) =\sin^2(5t)$ and $h_3(t) =\cos(4t)$ by numerical integration, then computing the output as,
 \begin{equation}
 \begin{split}
     y(t) = \min(5f_1(t)f_2(t)+5f^3_3(t), 1) + \epsilon,
 \end{split}
 \end{equation}
 with $\epsilon \sim \mathcal{N}(0, 0.05^2)$. We generate 1200 points
 in the range $t=[-20, 20]$ and use a random subset of a third for
 training and the rest for testing. Table \ref{toy_table} shows the
 normalised mean square errors (NMSEs) and negative log probability
 densities (NLPDs) on the test set for the NVKM with various values of
 $C$ as well as the GPCM, with repeats using a different random
 train/test split, and different random seeds.\footnote{Results
   generated using the implementation available at
   \hyperlink{https://github.com/wesselb/gpcm}{\texttt{github.com/wesselb/gpcm}}}
 As we would expect, the NMSE values are very similar for the NVKM
 with $C=1$ and the GPCM, since the models are nearly
 equivalent except for the prior on the input GP. Interestingly the
 NLPD values are better for the GPCM than the NVKM with
 $C=1$, likely due to the fact we do not optimise the noise jointly
 with the bound. As $C$ increases the performance of the NVKM improves
 until $C=4$. The fact performance does not improve after $C=3$ illustrates the difficulty of identifying higher order nonlinearities in a relatively small training set, an effect supported by the results of the \textit{Cascaded Tanks} experiment in the
   following section. Although the $C=4$ model does have more
 capacity to represent nonlinearities, the optimisation procedure is
 challenging, illustrated by the high variance of the results. Plots of the predictions
 for the model can be seen in Figure
 \ref{toy_figure}. We can see that increasing the non-linearity for
 the NVKMs allows the sharp spike and the finer grained features, as
 well as the hard nonlinearities, to be captured
 simultaneously.
  \begin{wraptable}{R}{6.5cm}
  \vspace{-8mm}
  \caption{Comparison of performance on the \textit{Cascaded Tanks} dataset, with the last four models reported in \citep{mattos2017recurrent}. $H$ indicates the number of hidden layers in the RGP.}
  \label{tanks_table}
  \centering
  \begin{tabular}{lll}
    \toprule
    Model     & RMSE     & NLPD\\
    \midrule
    IO-NVKM ($C=1$) & 0.835	& 1.724   \\
    IO-NVKM ($C=2$) & 0.716	& 1.311  \\
    IO-NVKM ($C=3$) & 0.532	& \textbf{0.879}  \\
    IO-NVKM ($C=4$) & 0.600	& 0.998\\
    RGP ($H=1$) &  0.797	& 2.33 \\
    RGP ($H=2$) & \textbf{0.308}	&  7.79 \\
    GP-NARX & 1.50 & 1080   \\
    Var. GP-NARX & 0.504	& 119.3   \\
    \bottomrule
      \vspace{-10mm}
  \end{tabular}
\end{wraptable} 
 \subsection{Cascaded tanks}\label{tanks_sec}

\begin{figure}
    \includegraphics[width=\textwidth]{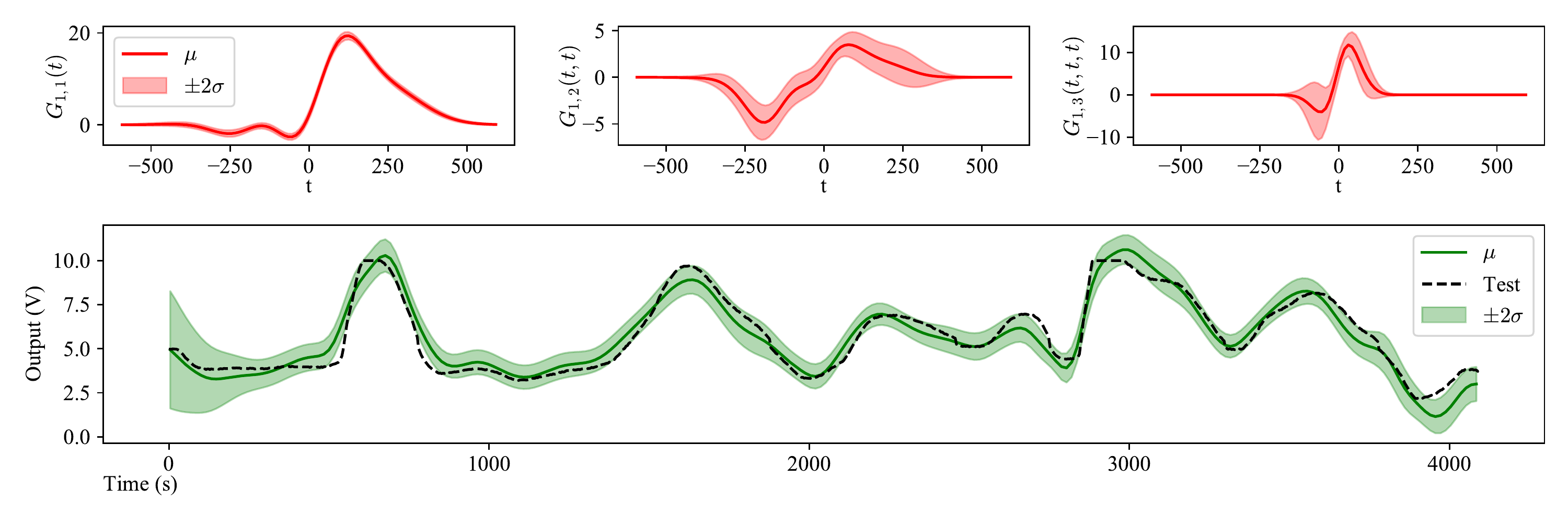}
  \caption{(Top) Diagonal of the inferred Volterra kernels
      for the the IO-NVKM with $C=3$, showing $2\sigma$ confidence region. (Bottom) The predicted output for the test set, with the dashed line showing the true values.}\label{tanks_figure}
\end{figure}

To demonstrate the IO-NVKM , we use a standard benchmark for
non-linear systems identification know as \textit{Cascaded Tanks}
\citep{schoukens2017three}.\footnote{Available at
  \hyperlink{https://sites.google.com/view/nonlinear-benchmark/}{\texttt{sites.google.com/view/nonlinear-benchmark/}}}
The system comprises two vertically stacked tanks filled with water,
with water being pumped from a reservoir to the top tank, which then
drains into the lower tank and finally back to the reservoir. The
training data is two time series of 1024 points, one being the input
to the system, which is the voltage fed into the pump, and the second
being the output, which is the measured water level in the lower
tank. For testing, an additional input signal,
again of 1024 points, is provided, and the task is to predict the
corresponding output water level. The system is considered
challenging because it contains hard nonlinearities when the tanks
reach maximum capacity and overflow (see the regions around 600s and
2900s in Figure \ref{tanks_figure}), it has unobserved internal state,
and has a relatively small training set. Table \ref{tanks_table} shows
the predictive root mean square errors (RMSEs) and NLPDs for the
IO-NVKM with various $C$, as well as four other GP based models for
system identification from \citep{mattos2017recurrent}. For each $C$,
five random settings of VK ranges were tested, and each training
was repeated three times with different
initialisations. The setting and
initialisation with the lowest combined NLPD on the training input and
output data is shown. Although the RGP with $H=2$ provides the best RMSE of the
model, this comes at the cost of poor NLPD values. All IO-NVKMs achieve
considerably better NLPD values than the alternatives indicating much
better quantification of uncertainty. Of the IO-NVKMs, $C=3$ performs
best in both metrics. Figure \ref{tanks_figure} show the predictions
of the $C=3$ model on the test set, as well as the inferred VKs. The
uncertainty in the VKs increases with their order, which is natural
given the difficulty of estimating higher order nonlinear effects from
a small training set. It should be noted that \citet{WORDEN2018194}
achieve a much lower RMSE of 0.191 by using a specific physics model
of the system in tandem with a GP-NARX model, but since we are
considering purely data driven approaches here, it is not directly
comparable.
 \subsection{Weather data}\label{weather_sec}
 To illustrate the utility of the NVKM for multiple output regression
 problems, we consider a popular benchmark in MOGP literature,
 consisting of multiple correlated time series of air temperature
 measurements taken at four nearby locations on the south coast of
 England, originally described by \citet{nguyen2014collaborative},
 which we refer to as \textit{Weather}.\footnote{Available for
   download in a convenient from using the \texttt{wbml} package,
   \hyperlink{https://github.com/wesselb/wbml}{\texttt{github.com/wesselb/wbml}}}
 The four series are named Bramblemet, Sotonmet, Cambermet and Chimet,
 with 1425, 1097, 1441, and 1436 data points, respectively. Bramblemet
 and Sotonmet both contain regions of truly missing data, 173 and 201
 points in a continuous region are artificially removed form Cambermet
 and Chimet with the task being to predict them based on the all the
 other data.
\begin{figure}
    \includegraphics[width=1.0\textwidth]{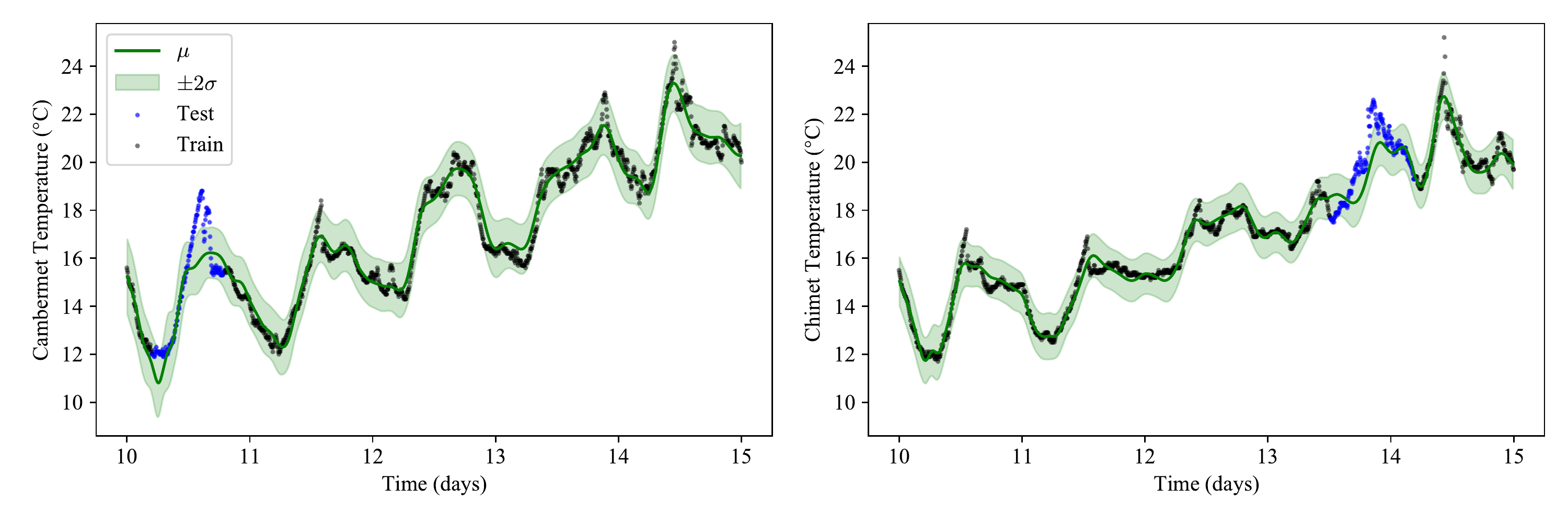}
  \caption{Predictive means and $2\sigma$ confidence regions from the NVKM with $C=3$, on the Cambermet and Chimet outputs from the \textit{Weather} data set. Blue points are the artificially removed test data, black points are training data.}\label{weather_figure}
    \end{figure}
\begin{table}
  \caption{Comparison of performance on the \textit{Weather} data set, for the NVKM mean and standard deviation of three initialisation is shown, along with the best model from \citep{alvarez2019non}}
  \label{weather_table}
  \centering
\begin{tabular}{
  lllll
  S[table-format=8.0(7)]
  }
    \toprule
    \multicolumn{1}{c}{} &\multicolumn{2}{c}{Cambermet} &\multicolumn{2}{c}{Chimet}                   \\
    \cmidrule(r){2-5} 
    Model     & NMSE     & NLPD & NMSE     & NLPD \\
    \midrule
    NVKM ($C=1$) &    \textbf{0.212\textpm	0.085} & 	\textbf{2.182\textpm	0.743} & 	1.669\textpm	0.052 & 	7.148\textpm	0.111 \\
    NVKM ($C=2$) & 	 0.440\textpm	0.286 & 	3.884\textpm	2.380 & 	0.939\textpm	0.216 & 	4.143\textpm	1.197\\
    NVKM ($C=3$) &   0.253\textpm	0.002 	&  2.390\textpm	0.123 &	 0.871\textpm	0.394 &	 3.994\textpm	1.924   \\
    NCMOGP ($C=3$)&  0.44 	&  2.33  &  \textbf{0.43} 	&  \textbf{2.18}  \\
    \bottomrule
  \end{tabular}
\end{table}
Table \ref{weather_table} shows the performance of the multiple output
NVKM on the \textit{Weather} dataset, along with the best performing
NCMOGP model of \citet{alvarez2019non}. For each $C$, five random
settings of VK ranges were tested, with each training being repeated four
times with different initialisations, the setting with the best
average NLPD value on the training data is shown. All NVKM models show
better or equivalent performance than the NCMOGP on the Cambermet
output, but all show worse performance on the Chimet output, although
on the Chimet output the variance between repeats is high. It should be noted that
the LFM reported by \citet{guarnizo2018fast} achieves much lower
scores, having NMSEs of $0.11$ and $0.19$ on Cambermet and Chimet
respectively, but that model uses six latent functions as opposed to a
single latent function for the NVKM and NCMOGP. Including multiple
latent functions may lead to large performance improvements for the
NVKM and is a promising direction for future work.
\section{Discussion}\label{discussion}
\paragraph{Societal Impacts} Accurate methods for system identification are key to the functioning of modern aircraft \citep{morelli2016aircraft}, this includes military aircraft, and specifically unmanned aerial vehicles equipped with weapons. It is possible that improved models for system identification could lead such aircraft to be more effective, and thus more deadly. GP and MOGP models have long been applied to problems in robotics \citep{deisenroth2013gaussian,williams208multi}. Better inclusion of nonlinearities in these models may enhance the ability of robots, potentially leading to loss of jobs and livelihoods to automation.
\paragraph{Future Work} The are a number of extensions to both the
NVKM and IO-NVKM that could lead to substantial improvements in
performance. As briefly mentioned in Section \ref{experiements}, the
number of inducing points required for the VKs scales exponentially
with the order of the series, meaning it is difficult to represent
complex features in the higher order terms, without using a
computationally intractable number of points. Whilst initially we saw
the increased flexibility of non-separable VKs as a virtue, it may be
that introducing separability leads to more powerful models, since the
number of points needed to specify separable VKs scales
linearly. Currently the models do not support multidimensional inputs,
but this could be easily added, requiring the computation of a few extra
integrals. For the multiple output model, allowing a shared set of
latent functions, with the input to each output's Volterra series being a trainable
linear combination, in a similar way to LFMs, is highly likely to
improve performance especially for problems with a large number of
outputs.

\paragraph{Conclusions} We have presented a novel model which
  uses Gaussian processes to learn the kernels of the Volterra series
  non-parametrically, allowing for the effective modeling of data with
  nonlinear properties. We have developed fast and scalable sampling
  and inference methods for the the model and show its performance on single and multiple output regression problems. Additionally, a modification to the model was presented that achieves significantly better uncertainty quantification than competitors on a challenging benchmark for nonlinear systems identification. 
  
\section*{Acknowledgements}
We thank Lizzy Cross, Felipe Tobar, and Carl Henrik Ek for helpful and insightful conversations. We would also like to thank Wessel Bruinsma for useful discussions and advice, as well as support in using the code for the GPCM. Magnus Ross and  Michael T. Smith thank the Department of Computer Science at the University of Sheffield financial support. Mauricio A. \'Alvarez has been financed by the EPSRC Research
  Projects EP/R034303/1, EP/T00343X/2 and EP/V029045/1.

{\footnotesize
\bibliography{refs}}

\begin{thebibliography}{37}
\providecommand{\natexlab}[1]{#1}
\providecommand{\url}[1]{\texttt{#1}}
\expandafter\ifx\csname urlstyle\endcsname\relax
  \providecommand{\doi}[1]{doi: #1}\else
  \providecommand{\doi}{doi: \begingroup \urlstyle{rm}\Url}\fi

\bibitem[Alvarez et~al.(2009)Alvarez, Luengo, and Lawrence]{alvarez2009latent}
M.~Alvarez, D.~Luengo, and N.~D. Lawrence.
\newblock Latent force models.
\newblock In \emph{Artificial Intelligence and Statistics}, pages 9--16. PMLR,
  2009.

\bibitem[\'Alvarez et~al.(2019)\'Alvarez, Ward, and Guarnizo]{alvarez2019non}
M.~\'Alvarez, W.~Ward, and C.~Guarnizo.
\newblock Non-linear process convolutions for multi-output {G}aussian
  processes.
\newblock pages 1969--1977, 2019.

\bibitem[\'{A}lvarez et~al.(2012)\'{A}lvarez, Rosasco, and
  Lawrence]{alvarez2012kernels}
M.~A. \'{A}lvarez, L.~Rosasco, and N.~D. Lawrence.
\newblock Kernels for vector-valued functions: A review.
\newblock \emph{Found. Trends Mach. Learn.}, 4\penalty0 (3):\penalty0
  195–266, Mar. 2012.
\newblock ISSN 1935-8237.
\newblock \doi{10.1561/2200000036}.
\newblock URL \url{https://doi.org/10.1561/2200000036}.

\bibitem[Barry and {Ver Hoef}(1996)]{Barry:blackbox96}
R.~P. Barry and J.~M. {Ver Hoef}.
\newblock Blackbox kriging: spatial prediction without specifying variogram
  models.
\newblock \emph{Journal of Agricultural, Biological and Environmental
  Statistics}, 1\penalty0 (3):\penalty0 297--322, 1996.

\bibitem[Benton et~al.(2019)Benton, Maddox, Salkey, Albinati, and
  Wilson]{benton2019function}
G.~Benton, W.~J. Maddox, J.~Salkey, J.~Albinati, and A.~G. Wilson.
\newblock Function-space distributions over kernels.
\newblock In \emph{Advances in Neural Information Processing Systems}, pages
  14965--14976, 2019.

\bibitem[Bradbury et~al.(2018)Bradbury, Frostig, Hawkins, Johnson, Leary,
  Maclaurin, Necula, Paszke, Vander{P}las, Wanderman-{M}ilne, and
  Zhang]{jax2018github}
J.~Bradbury, R.~Frostig, P.~Hawkins, M.~J. Johnson, C.~Leary, D.~Maclaurin,
  G.~Necula, A.~Paszke, J.~Vander{P}las, S.~Wanderman-{M}ilne, and Q.~Zhang.
\newblock {JAX}: composable transformations of {P}ython+{N}um{P}y programs,
  2018.
\newblock URL \url{http://github.com/google/jax}.

\bibitem[Bruinsma(2016)]{bruinsma2016generalised}
W.~Bruinsma.
\newblock \emph{The Generalised {G}aussian Process Convolution Model}.
\newblock {MPhil.} thesis, University of Cambridge, 2016.

\bibitem[Cheng et~al.(2017)Cheng, Peng, Zhang, and Meng]{cheng2017volterra}
C.~Cheng, Z.~Peng, W.~Zhang, and G.~Meng.
\newblock Volterra-series-based nonlinear system modeling and its engineering
  applications: A state-of-the-art review.
\newblock \emph{Mechanical Systems and Signal Processing}, 87:\penalty0 340 --
  364, 2017.
\newblock ISSN 0888-3270.
\newblock \doi{https://doi.org/10.1016/j.ymssp.2016.10.029}.
\newblock URL
  \url{http://www.sciencedirect.com/science/article/pii/S0888327016304393}.

\bibitem[Deisenroth et~al.(2013)Deisenroth, Fox, and
  Rasmussen]{deisenroth2013gaussian}
M.~P. Deisenroth, D.~Fox, and C.~E. Rasmussen.
\newblock {G}aussian processes for data-efficient learning in robotics and
  control.
\newblock \emph{IEEE transactions on pattern analysis and machine
  intelligence}, 37\penalty0 (2):\penalty0 408--423, 2013.

\bibitem[Duvenaud et~al.(2011)Duvenaud, Nickisch, and
  Rasmussen]{duvenaud2011additive}
D.~K. Duvenaud, H.~Nickisch, and C.~Rasmussen.
\newblock Additive {G}aussian processes.
\newblock In J.~Shawe-Taylor, R.~Zemel, P.~Bartlett, F.~Pereira, and K.~Q.
  Weinberger, editors, \emph{Advances in Neural Information Processing
  Systems}, volume~24. Curran Associates, Inc., 2011.
\newblock URL
  \url{https://proceedings.neurips.cc/paper/2011/file/4c5bde74a8f110656874902f07378009-Paper.pdf}.

\bibitem[Guarnizo and {\'A}lvarez(2018)]{guarnizo2018fast}
C.~Guarnizo and M.~A. {\'A}lvarez.
\newblock Fast kernel approximations for latent force models and convolved
  multiple-output {G}aussian processes.
\newblock In \emph{Uncertainty in Artificial Intelligence: Proceedings of the
  Thirty-Fourth Conference (2018)}. AUAI Press, 2018.

\bibitem[Hartikainen et~al.(2012)Hartikainen, Sepp{\"a}nen, and
  S{\"a}rkk{\"a}]{hartikainen2012state}
J.~Hartikainen, M.~Sepp{\"a}nen, and S.~S{\"a}rkk{\"a}.
\newblock State-space inference for non-linear latent force models with
  application to satellite orbit prediction.
\newblock In \emph{Proceedings of the 29th International Coference on
  International Conference on Machine Learning}, pages 723--730, 2012.

\bibitem[Higdon(2002)]{higdon2002space}
D.~Higdon.
\newblock Space and space-time modeling using process convolution.
\newblock In \emph{Quantitative methods for current environmental issues},
  pages 37--56. Springer, 2002.

\bibitem[Hoffman et~al.(2013)Hoffman, Blei, Wang, and
  Paisley]{hoffman2013stochastic}
M.~D. Hoffman, D.~M. Blei, C.~Wang, and J.~Paisley.
\newblock Stochastic variational inference.
\newblock \emph{Journal of Machine Learning Research}, 14\penalty0 (5), 2013.

\bibitem[Kingma and Ba(2015)]{DBLP:journals/corr/KingmaB14}
D.~P. Kingma and J.~Ba.
\newblock Adam: {A} method for stochastic optimization.
\newblock In Y.~Bengio and Y.~LeCun, editors, \emph{3rd International
  Conference on Learning Representations, {ICLR} 2015, San Diego, CA, USA, May
  7-9, 2015, Conference Track Proceedings}, 2015.
\newblock URL \url{http://arxiv.org/abs/1412.6980}.

\bibitem[Kocijan et~al.(2005)Kocijan, Girard, Banko, and
  Murray-Smith]{kocijan2005dynamic}
J.~Kocijan, A.~Girard, B.~Banko, and R.~Murray-Smith.
\newblock Dynamic systems identification with {G}aussian processes.
\newblock \emph{Mathematical and Computer Modelling of Dynamical Systems},
  11\penalty0 (4):\penalty0 411--424, 2005.
\newblock \doi{10.1080/13873950500068567}.
\newblock URL \url{https://doi.org/10.1080/13873950500068567}.

\bibitem[Lawrence et~al.(2009)Lawrence, Rattray, and
  Titsias]{NIPS2008_5487315b}
N.~Lawrence, M.~Rattray, and M.~Titsias.
\newblock Efficient sampling for {G}aussian process inference using control
  variables.
\newblock In D.~Koller, D.~Schuurmans, Y.~Bengio, and L.~Bottou, editors,
  \emph{Advances in Neural Information Processing Systems}, volume~21. Curran
  Associates, Inc., 2009.
\newblock URL
  \url{https://proceedings.neurips.cc/paper/2008/file/5487315b1286f907165907aa8fc96619-Paper.pdf}.

\bibitem[Mattos(2017)]{mattos2017recurrent}
C.~L.~C. Mattos.
\newblock \emph{Recurrent {G}aussian processes and robust dynamical modeling}.
\newblock PhD thesis, 2017.

\bibitem[Mattos et~al.(2016)Mattos, Dai, Damianou, Forth, Barreto, and
  Lawrence]{Mattos-recurrent16}
C.~L.~C. Mattos, Z.~Dai, A.~Damianou, J.~Forth, G.~A. Barreto, and N.~D.
  Lawrence.
\newblock Recurrent {G}aussian processes.
\newblock In H.~Larochelle, B.~Kingsbury, and S.~Bengio, editors,
  \emph{Proceedings of the International Conference on Learning
  Representations}, volume~3, Caribe Hotel, San Juan, PR, 2016.
\newblock URL
  \url{http://inverseprobability.com/publications/mattos-recurrent16.html}.

\bibitem[Morelli and Klein(2016)]{morelli2016aircraft}
E.~A. Morelli and V.~Klein.
\newblock \emph{Aircraft system identification: theory and practice}, volume~2.
\newblock Sunflyte Enterprises Williamsburg, VA, 2016.

\bibitem[Nguyen et~al.(2014)Nguyen, Bonilla, et~al.]{nguyen2014collaborative}
T.~V. Nguyen, E.~V. Bonilla, et~al.
\newblock Collaborative multi-output {G}aussian processes.
\newblock In \emph{UAI}, pages 643--652. Citeseer, 2014.

\bibitem[Rahimi and Recht(2007)]{rahami2007random}
A.~Rahimi and B.~Recht.
\newblock Random features for large-scale kernel machines.
\newblock \emph{NIPS}, 3, 2007.

\bibitem[Rasmussen and Williams(2005{\natexlab{a}})]{gpmlbook}
C.~E. Rasmussen and C.~K.~I. Williams.
\newblock \emph{{G}aussian Processes for Machine Learning (Adaptive Computation
  and Machine Learning)}.
\newblock The MIT Press, 2005{\natexlab{a}}.
\newblock ISBN 026218253X.

\bibitem[Rasmussen and Williams(2005{\natexlab{b}})]{gpmlbookpg203}
C.~E. Rasmussen and C.~K.~I. Williams.
\newblock \emph{{G}aussian Processes for Machine Learning (Adaptive Computation
  and Machine Learning)}, pages 203--204.
\newblock The MIT Press, 2005{\natexlab{b}}.
\newblock ISBN 026218253X.

\bibitem[Schoukens and No{\"e}l(2017)]{schoukens2017three}
M.~Schoukens and J.~P. No{\"e}l.
\newblock Three benchmarks addressing open challenges in nonlinear system
  identification.
\newblock \emph{IFAC-PapersOnLine}, 50\penalty0 (1):\penalty0 446--451, 2017.

\bibitem[Svensson and Sch{\"o}n(2017)]{svensson2017flexible}
A.~Svensson and T.~B. Sch{\"o}n.
\newblock A flexible state--space model for learning nonlinear dynamical
  systems.
\newblock \emph{Automatica}, 80:\penalty0 189--199, 2017.

\bibitem[Svensson et~al.(2016)Svensson, Solin, Särkkä, and
  Schön]{svensson16compuatoinally}
A.~Svensson, A.~Solin, S.~Särkkä, and T.~Schön.
\newblock Computationally efficient bayesian learning of {G}aussian process
  state space models.
\newblock In A.~Gretton and C.~C. Robert, editors, \emph{Proceedings of the
  19th International Conference on Artificial Intelligence and Statistics},
  volume~51 of \emph{Proceedings of Machine Learning Research}, pages 213--221,
  Cadiz, Spain, 09--11 May 2016. PMLR.
\newblock URL \url{http://proceedings.mlr.press/v51/svensson16.html}.

\bibitem[Titsias(2009)]{titsias2009variational}
M.~Titsias.
\newblock Variational learning of inducing variables in sparse {G}aussian
  processes.
\newblock In \emph{Artificial Intelligence and Statistics}, pages 567--574,
  2009.

\bibitem[Titsias and L{\'a}zaro-Gredilla(2014)]{titsias2014doubly}
M.~Titsias and M.~L{\'a}zaro-Gredilla.
\newblock Doubly stochastic variational bayes for non-conjugate inference.
\newblock In \emph{International conference on machine learning}, pages
  1971--1979. PMLR, 2014.

\bibitem[Tobar et~al.(2015)Tobar, Bui, and Turner]{tobar2015learning}
F.~Tobar, T.~D. Bui, and R.~E. Turner.
\newblock Learning stationary time series using {G}aussian processes with
  nonparametric kernels.
\newblock In \emph{Advances in Neural Information Processing Systems}, pages
  3501--3509, 2015.

\bibitem[{Ver Hoef} and Barry(1998)]{verHoef:convolution98}
J.~M. {Ver Hoef} and R.~P. Barry.
\newblock Constructing and fitting models for cokriging and multivariable
  spatial prediction.
\newblock \emph{Journal of Statistical Plannig and Inference}, 69:\penalty0
  275--294, 1998.

\bibitem[Ward et~al.(2020)Ward, Ryder, Prangle, and Alvarez]{ward2020black}
W.~Ward, T.~Ryder, D.~Prangle, and M.~Alvarez.
\newblock Black-box inference for non-linear latent force models.
\newblock In \emph{International Conference on Artificial Intelligence and
  Statistics}, pages 3088--3098. PMLR, 2020.

\bibitem[Williams et~al.(2009)Williams, Klanke, Vijayakumar, and
  Chai]{williams208multi}
C.~Williams, S.~Klanke, S.~Vijayakumar, and K.~Chai.
\newblock Multi-task {G}aussian process learning of robot inverse dynamics.
\newblock In D.~Koller, D.~Schuurmans, Y.~Bengio, and L.~Bottou, editors,
  \emph{Advances in Neural Information Processing Systems}, volume~21. Curran
  Associates, Inc., 2009.
\newblock URL
  \url{https://proceedings.neurips.cc/paper/2008/file/15d4e891d784977cacbfcbb00c48f133-Paper.pdf}.

\bibitem[Wilson and Adams(2013)]{wilson2013gaussian}
A.~Wilson and R.~Adams.
\newblock {G}aussian process kernels for pattern discovery and extrapolation.
\newblock In \emph{International conference on machine learning}, pages
  1067--1075, 2013.

\bibitem[Wilson et~al.(2016)Wilson, Hu, Salakhutdinov, and
  Xing]{wilson2016deep}
A.~G. Wilson, Z.~Hu, R.~Salakhutdinov, and E.~P. Xing.
\newblock Deep kernel learning.
\newblock In \emph{Artificial intelligence and statistics}, pages 370--378,
  2016.

\bibitem[Wilson et~al.(2020)Wilson, Borovitskiy, Terenin, Mostowsky, and
  Deisenroth]{wilson2020efficiently}
J.~Wilson, V.~Borovitskiy, A.~Terenin, P.~Mostowsky, and M.~Deisenroth.
\newblock Efficiently sampling functions from {G}aussian process posteriors.
\newblock In H.~D. III and A.~Singh, editors, \emph{Proceedings of the 37th
  International Conference on Machine Learning}, volume 119 of
  \emph{Proceedings of Machine Learning Research}, pages 10292--10302. PMLR,
  13--18 Jul 2020.
\newblock URL \url{http://proceedings.mlr.press/v119/wilson20a.html}.

\bibitem[Worden et~al.(2018)Worden, Barthorpe, Cross, Dervilis, Holmes, Manson,
  and Rogers]{WORDEN2018194}
K.~Worden, R.~Barthorpe, E.~Cross, N.~Dervilis, G.~Holmes, G.~Manson, and
  T.~Rogers.
\newblock On evolutionary system identification with applications to nonlinear
  benchmarks.
\newblock \emph{Mechanical Systems and Signal Processing}, 112:\penalty0
  194--232, 2018.
\newblock ISSN 0888-3270.
\newblock \doi{https://doi.org/10.1016/j.ymssp.2018.04.001}.
\newblock URL
  \url{https://www.sciencedirect.com/science/article/pii/S0888327018301912}.

\end{thebibliography}

\appendix
\section*{Appendix}
\section{Derivation of explicit sampling equations}\label{explder}
Recall that we wish to analytically compute the integral in Equation (5), which was given by 
\begin{equation*}
    (f_d|\{ \mathbf{v}^G_{d,c}\}^{C}_{c=1} , \mathbf{v}^u)(t) = \sum_{c=1}^{C} \int^\infty_{-\infty}e^{-\alpha \sum^c_{i=1} (t - \tau_i)^2} (G'_{d, c}|\mathbf{v}^G_{d,c})(t-\tau_1, \ldots, t-\tau_c) \prod_{j=1}^{c} (u|\mathbf{v}^{u})(\tau_{j}) \text{d}\tau_j,
    \label{nvkms}
\end{equation*}
with all the processes $\{G_{d,c}\}^{C, D}_{c, d=1}$ and $u$ having the SE covariances, given by $k_{SE}(t, t') = \sigma^2 \exp(-p\|t-t'\|^2)$, where $\sigma$ is the amplitude of the process,  and $p$ is the precision, which is related to the length scale $l$ by $p=\frac{1}{2l^2}$. For notational simplicity we will compute the integral for the $c$th term, and drop the subscripts on $G$, we then substitute Equation (4) for $G$, giving, 
\begin{equation*}
    \begin{split}
    I_{c,d} =&\int^{\infty}_{-\infty} e^{-\alpha \sum^c_{m=1} (t - \tau_m)^2} (G' | \mathbf{v}^{G})(t-\tau_1, \ldots, t-\tau_c) \prod_{k=1}^{c} (u|\mathbf{v}^{u})(\tau_{k}) \text{d} \tau_{k} \\
    =&\int^{\infty}_{-\infty} e^{-\alpha \sum^c_{m=1} (t - \tau_m)^2} \Big(\sum^{N_b}_{i=1} w^{(G)}_i\phi^{(G)}_i(t -\boldsymbol{\tau}) + \sum^{M^{(G)}}_{j=1}q^{(G)}_j k^{(G)}(t -\boldsymbol{\tau}, \mathbf{z}^{(G)}_j) \Big)\prod_{k=1}^{c} (u|\mathbf{v}^{u})(\tau_{k}) \text{d} \tau_{k} \\
    =& \sum^{N_b}_{i=1} w^{(G)}_i\underbrace{ \int^{\infty}_{-\infty} e^{-\alpha \sum^c_{m=1} (t - \tau_m)^2} \phi^{(G)}_i(t -\boldsymbol{\tau})\prod_{k=1}^{c} (u|\mathbf{v}^{u})(\tau_{k}) \text{d} \tau_{k}}_{I_1}\\
    +&\sum^{M^{(G)}}_{j=1}q^{(G)}_j \underbrace{\int^{\infty}_{-\infty} e^{-\alpha \sum^c_{m=1} (t - \tau_m)^2} k^{(G)}(t -\boldsymbol{\tau}, \mathbf{z}^{(G)}_j) \prod_{k=1}^{c} (u|\mathbf{v}^{u})(\tau_{k}) \text{d} \tau_{k}}_{I_2}
    \end{split}
\end{equation*}
with $\boldsymbol{\tau}= (\tau_1, \dots, \tau_c)$ and $t-\boldsymbol{\tau} = (t - \tau_1, \dots, t - \tau_c)$, with superscripts indicating which of the VK or input process the symbols from Equation (4) are associated with. We can see that there are two separate integrals to deal with, $I_1$ and $I_2$.

\subsection{$I_1$}
We have
\begin{equation*}
\begin{split}
   I_1 =&\int^{\infty}_{-\infty} e^{-\alpha \sum^{c}_{m=1} (t - \tau_m)^2} \phi^{(G)}_i(t -\boldsymbol{\tau})\prod_{k=1}^{c} (u|\mathbf{v}^{u})(\tau_{k}) \text{d} \tau_{k} \\
   =&\sqrt{\frac{2}{N_b}}\int^{\infty}_{-\infty} e^{-\alpha \sum^{c}_{m=1} (t - \tau_m)^2} \cos( \sum^c_{j=1} \theta_{i,j}(t-\tau_j) +\beta_i)\prod_{k=1}^{c} (u|\mathbf{v}^{u})(\tau_{k}) \text{d} \tau_{k}\\
   =& \frac{1}{2}\sqrt{\frac{2}{N_b}}\int^{\infty}_{-\infty}e^{-\alpha \sum^{c}_{m=1} (t - \tau_m)^2} e^{i(\sum^c_{j=1} \theta_{i,j}(t-\tau_j) +\beta_i)}\prod_{k=1}^{c} (u|\mathbf{v}^{u})(\tau_{k}) \text{d} \tau_{k} \\
   +& \frac{1}{2}\sqrt{\frac{2}{N_b}}\int^{\infty}_{-\infty}e^{-\alpha \sum^{c}_{m=1} (t - \tau_m)^2} e^{-i(\sum^c_{j=1} \theta_{i,j}(t-\tau_j) +\beta_i)}\prod_{k=1}^{c} (u|\mathbf{v}^{u})(\tau_{k}) \text{d} \tau_{k} \\
   =& \frac{1}{2}\sqrt{\frac{2}{N_b}}e^{i\beta_i}\int^{\infty}_{-\infty} \prod^c_{j=1}e^{-\alpha (t - \tau_j)^2 +i\theta_{i,j}(t-\tau_j)}\prod_{k=1}^{c} (u|\mathbf{v}^{u})(\tau_{k}) \text{d} \tau_{k} \\
   +& \frac{1}{2}\sqrt{\frac{2}{N_b}}e^{-i\beta_i}\int^{\infty}_{-\infty} \prod^c_{j=1}e^{-\alpha (t - \tau_j)^2 -i\theta_{i,j}(t-\tau_j)}\prod_{k=1}^{c} (u|\mathbf{v}^{u})(\tau_{k}) \text{d} \tau_{k} \\
      =& \frac{1}{2}\sqrt{\frac{2}{N_b}}e^{i\beta_i}\prod^c_{j=1} \int^{\infty}_{-\infty} e^{-\alpha (t - \tau)^2 +i\theta_{i,j}(t-\tau)}  (u|\mathbf{v}^{u})(\tau) \text{d} \tau  \\
   +& \frac{1}{2}\sqrt{\frac{2}{N_b}}e^{-i\beta_i}\prod^c_{j=1} \int^{\infty}_{-\infty} e^{-\alpha (t - \tau)^2 -i\theta_{i,j}(t-\tau) }  (u|\mathbf{v}^{u})(\tau) \text{d} \tau, \\
   \end{split}
\end{equation*}
so we need to evaluate 
\begin{equation*}
    \begin{split}
&\int^{\infty}_{-\infty} e^{-\alpha (t - \tau)^2 \pm i\theta^{(G)}_{i,j}(t-\tau)}  (u|\mathbf{v}^{u})(\tau) \text{d} \tau \\
=& \int^{\infty}_{-\infty} e^{-\alpha (t - \tau)^2 \pm i\theta^{(G)}_{i,j}(t-\tau)}  \Big(\sum^{N_b}_{m=1} w^{(u)}_m\phi^{(u)}_m(\tau) + \sum^{M^{(u)}}_{n=1}q^{(u)}_n k^{(u)}(\tau, z^{(u)}_n) \Big)\text{d} \tau \\
=&\sqrt{\frac{2}{N_b}}\sum^{N_b}_{m=1}  w^{(u)}_m \int^{\infty}_{-\infty} e^{-\alpha (t - \tau)^2 \pm i\theta^{(G)}_{i,j}(t-\tau) } \cos(\theta^{(u)}_m \tau + \beta^{(u)}_m)d\tau \\
+& \sigma_u^2\sum^{M^{(u)}}_{n=1}q^{(u)}_n \int^{\infty}_{-\infty} e^{-\alpha (t - \tau)^2 \pm i\theta^{(G)}_{i,j}(t-\tau) } e^{-p_u(\tau - z^{(u)}_n)^2}d\tau \\
=& \sqrt{\frac{2}{N_b}}\sum^{N_b}_{m=1}  w^{(u)}_m I_{1a}(t; \alpha, \pm \theta^{(G)}_{i,j}, \theta^{(u)}_m, \beta^{(u)}_m) + \sigma_u^2\sum^{M^{(u)}}_{n=1}q^{(u)}_n I_{1b}(t; \alpha, \pm\theta^{(G)}_{i,j}, p_u, z^{(u)}_n). 
    \end{split}
\end{equation*}
We can find explicit forms for $I_{1a}$, 
\begin{equation*}
    \begin{split}
    I_{1a}(t; \alpha, \theta_1, \theta_2, \beta_2) &=   \int^{\infty}_{-\infty} e^{-\alpha (t - \tau)^2 + i   \theta_1(t-\tau)} \cos(\theta_2\tau + \beta_2)d\tau \\
    &= \frac{\sqrt{\pi }}{2 \sqrt{\alpha}} \left(1+e^{\frac{\theta_1 \theta_2}{\alpha}+2 i \beta_2+2 i  \theta_2 t}  \right) e^{-\frac{(\theta_1+\theta_2)^2}{4 \alpha}-i (\beta_2+ \theta_2 t)},
    \end{split}
\end{equation*}
as well as for $I_{1b}$, 
\begin{equation*}
    \begin{split}
       I_{1b}(t; \alpha,  \theta_1, p_2, z_2) &= \int^{\infty}_{-\infty} e^{-\alpha (t - \tau)^2 + i\theta_1(t-\tau)} e^{-p_2(\tau - z_2)^2}d\tau \\
       &= \frac{\sqrt{\pi }}{\sqrt{\alpha+p_2}}  e^{\frac{ -4 \alpha p_2 (t-z_2)^2+i \theta_1 (4 p_2 t-4 p_2 z_2+i \theta_1)}{4 (\alpha+p_2)}}.
    \end{split}
\end{equation*}
Putting it all together we get that $I_1$ is given by,
\begin{equation*}
\begin{split}
    I_1 = \frac{1}{2}\sqrt{\frac{2}{N_b}}\Big( e^{i\beta^{(G)}_i}&\prod^c_{j=1}\Big[ \sqrt{\frac{2}{N_b}}\sum^{N_b}_{m=1}  w^{(u)}_m I_{1a}(t; \alpha, \theta^{(G)}_{i,j},  \theta^{(u)}_m, \beta^{(u)}_m) \\
     + \sigma_u^2&\sum^{M^{(u)}}_{n=1}q^{(u)}_n I_{1b}(t; \alpha, \theta^{(G)}_{i,j},  p_u, z^{(u)}_n)\Big] \\
    + e^{-i\beta^{(G)}_i} & \prod^c_{j=1} \Big[ \sqrt{\frac{2}{N_b}}\sum^{N_b}_{m=1}  w^{(u)}_m I_{1a}(t; \alpha, -\theta^{(G)}_{i,j}, - \theta^{(u)}_m, \beta^{(u)}_m) \\
    + \sigma_u^2 &\sum^{M^{(u)}}_{n=1}q^{(u)}_n I_{1b}(t; \alpha, -\theta^{(G)}_{i,j}, - p_u, z^{(u)}_n)\Big] \Big) 
    \end{split}
\end{equation*}
\subsection{$I_2$}
$I_2$ is given by 
\begin{equation*}
\int^{\infty}_{-\infty} e^{-\alpha \sum^c_{m=1} (t - \tau_m)^2} k^{(G)}(t -\boldsymbol{\tau}, \mathbf{z}^{(G)}_j) \prod_{k=1}^{c} (u|\mathbf{v}^{u})(\tau_{k}) \text{d} \tau_{k}
\end{equation*}
when we substitute in our expression for the SE kernel we get that,
\begin{equation*}
\begin{split}
      I_2 &=  \sigma^2_{G}\int^{\infty}_{-\infty} e^{-\alpha \sum^c_{m=1} (t - \tau_m)^2} e^{-p_G\sum_l (t - \tau_l - z^{(G)}_{j,l})^2}  \prod_{k=1}^{c} (u|\mathbf{v}^{u})(\tau_{k}) \text{d} \tau_{k}  \\
      &=  \sigma^2_{G} \int^{\infty}_{-\infty} \prod^c_{m=1} e^{-\alpha (t - \tau_m)^2} \prod^c_{l=1} e^{-p_G (t - \tau_l - z^{(G)}_{j,l})^2}  \prod_{k=1}^{c} (u|\mathbf{v}^{u})(\tau_{k}) \text{d} \tau_{k} \\ 
              &= \sigma^2_{G} \prod^c_{i=1} \int^{\infty}_{-\infty} e^{-\alpha (t - \tau)^2}  e^{-p_G (t - \tau - z^{(G)}_{j,i})^2} (u|\mathbf{v}^{u})(\tau) \text{d} \tau .\\
    \end{split}
\end{equation*}
Substituting in the expression for the input process $u$, we get
\begin{equation*}
\begin{split}
    I_2 &= \sigma^2_{G}\prod^c_{i=1} \int^{\infty}_{-\infty} e^{-\alpha (t - \tau)^2}  e^{-p_G (t - \tau - z^{(G)}_{j,i})^2} \Big(\sum^{N_b}_{m=1} w^{(u)}_m\phi^{(u)}_m(\tau) + \sum^{M^{(u)}}_{n=1}q^{(u)}_n k^{(u)}(\tau, z^{(u)}_n) \Big)\text{d} \tau \\
    &=\sigma^2_{G}\prod^c_{i=1} \Big[\sqrt{\frac{2}{N_b}}\sum^{N_b}_{m=1}  w^{(u)}_m \int^{\infty}_{-\infty} e^{-\alpha (t - \tau)^2 -p_G (t - \tau - z^{(G)}_{j,i})^2} \cos(\theta^{(u)}_m \tau + \beta^{(u)}_m)d\tau \\
&+ \sigma_u^2\sum^{M^{(u)}}_{n=1}q^{(u)}_n \int^{\infty}_{-\infty} e^{-\alpha (t - \tau)^2 -p_G (t - \tau - z^{(G)}_{j,i})^2} e^{-p_u(\tau - z^{(u)}_n)^2}d\tau \Big]\\
&= \sigma^2_{G}\prod^c_{i=1}\Big[ \sqrt{\frac{2}{N_b}}\sum^{N_b}_{m=1}  w^{(u)}_m I_{2a}(t; \alpha, p_G, z^{(G)}_{j, i}, \theta^{(u)}_m, \beta^{(u)}_m)\\
&+ \sigma_u^2\sum^{M^{(u)}}_{n=1}q^{(u)}_n I_{2b}(t; \alpha, p_G, z^{(G)}_{j, i}, p_u, z^{(u)}_n) \Big]. \\
\end{split}
\end{equation*}
The explicit forms for $I_{2a}$ and $I_{2b}$ are given by,
\begin{equation*}
    \begin{split}
    I_{2a}(t; \alpha, p_1, z_1, \theta_2, \beta_2) &= \int^{\infty}_{-\infty} e^{-\alpha (t - \tau)^2 -p_1 (t - \tau - z_1)^2} \cos(\theta_2 \tau + \beta_2)d\tau \\
    &= \frac{\sqrt{\pi }}{\sqrt{a+p_1}} e^{-\frac{4 a p_1 z_1^2+\theta_2^2}{4 (a+p_1)}} \cos \left(\theta_2\left(t-\frac{p_1 z_1}{a+p_1}\right)+\beta_2\right) \\
    I_{2b}(t; \alpha, p_1, z_1, p_2, z_2)  &= \int^{\infty}_{-\infty} e^{-\alpha (t - \tau)^2 -p_1 (t - \tau - z_1)^2} e^{-p_2(\tau - z_2)^2}d\tau \\
    &= \frac{\sqrt{\pi }}{\sqrt{a+p_1+p_2}} e^{-\frac{a \left(p_1 z_1^2+p_2 (t-z_2)^2\right)+p_1 p_2 (-t+z_1+z_2)^2}{a+p_1+p_2}}. 
\end{split}
\end{equation*}
\subsection{Final expression}
Collating $I_1$ and $I_2$, we get that the final expression for an explicit sample from the $c$th order term is    
\begin{equation*}
    (f_d|\{ \mathbf{v}^G_{d,c}\}^{C}_{c=1} , \mathbf{v}^u)(t) = \sum_{c=1}^{C} I_{c,d},
\end{equation*}
with
\begin{equation*}
    \begin{split}
    I_{c, d} =& \sum^{N_b}_{i=1}w^{(G)}_i\Bigg(\frac{1}{2}\sqrt{\frac{2}{N_b}}\Big( e^{i\beta^{(G)}_i}\prod^c_{j=1}\Big[ \sqrt{\frac{2}{N_b}}\sum^{N_b}_{m=1}  w^{(u)}_m I_{1a}(t; \alpha, \theta^{(G)}_{i,j},  \theta^{(u)}_m, \beta^{(u)}_m) \\
     &+ \sigma_u^2\sum^{M^{(u)}}_{n=1}q^{(u)}_n I_{1b}(t; \alpha, \theta^{(G)}_{i,j},  p_u, z^{(u)}_n)\Big] \\
    + &e^{-i\beta^{(G)}_i}  \prod^c_{j=1} \Big[ \sqrt{\frac{2}{N_b}}\sum^{N_b}_{m=1}  w^{(u)}_m I_{1a}(t; \alpha, -\theta^{(G)}_{i,j}, - \theta^{(u)}_m, \beta^{(u)}_m) \\
    &+ \sigma_u^2 \sum^{M^{(u)}}_{n=1}q^{(u)}_n I_{1b}(t; \alpha, -\theta^{(G)}_{i,j}, - p_u, z^{(u)}_n)\Big] \Big) \Bigg)  \\
    +&\sum^{M^{(G)}}_{j=1}q^{(G)}_j \Bigg(\sigma^2_{G}\prod^c_{i=1} \Big[\sqrt{\frac{2}{N_b}}\sum^{N_b}_{m=1}  w^{(u)}_m I_{2a}(t; \alpha, p_G, z^{(G)}_{i,j}, \theta^{(u)}_m, \beta^{(u)}_m)\\
&+ \sigma_u^2\sum^{M^{(u)}}_{n=1}q^{(u)}_n I_{2b}(t; \alpha, p_G, z^{(G)}_{i,j}, p_u, z^{(u)}_n) \Big]\Bigg).
    \end{split}
\end{equation*}

\section{Derivation of variational lower bound}\label{boundder}
Recall that the true joint distribution is given by
\begin{multline}
    p(\{\mathbf{y}_d\}^{D}_{d=1},\{G_{d,c}, \mathbf{v}^G_{d,c}\}^{C, D}_{c, d=1} , u, \mathbf{v}^u) =\\\prod^{D, N_d}_{d,i=1}  p(y_{d,i}|f_d(t_{d,i}) )
\prod^{D,C}_{d,c=1} p(G_{d,c} | \mathbf{v}^G_{d,c})p(\mathbf{v}^G_{d,c})p(u | \mathbf{v}^u) p(\mathbf{v}^u),
\end{multline}
and the variational distribution by
\begin{equation*}
q(\{G_{d,c}, \mathbf{v}^G_{d,c}\}^{C, D}_{c, d=1} , u, \mathbf{v}^u) =  \prod^{D,C}_{d,c=1} p(G_{d,c} | \mathbf{v}^G_{d,c})q(\mathbf{v}^G_{d,c})p(u | \mathbf{v}^u)q(\mathbf{v}^u).
\end{equation*}
The variational lower bound is the KL divergence between the variational posterior and true posterior, given by
\begin{equation*}
    \begin{split}
        \mathcal{F} = \int q(\{G_{d,c}, \mathbf{v}^G_{d,c}\}^{C, D}_{c, d=1} , u, \mathbf{v}^u) \log \frac{    p(\{\mathbf{y}_d\}^{D}_{d=1},\{G_{d,c}, \mathbf{v}^G_{d,c}\}^{C, D}_{c, d=1} , u, \mathbf{v}^u)}{q(\{G_{d,c}, \mathbf{v}^G_{d,c}\}^{C, D}_{c, d=1} , u, \mathbf{v}^u)} dS,
    \end{split}
\end{equation*}
where $dS$ represents the integral over all inducing points, all VK GPs and input GP. Substituting the expressions for each distribution we obtain,
\begin{equation*}
    \begin{split}
        \mathcal{F} &= \int q(\{G_{d,c}, \mathbf{v}^G_{d,c}\}^{C, D}_{c, d=1} , u, \mathbf{v}^u)  \\ 
        & \hspace{1cm}\times \log \frac{\prod^{D, N_d}_{d,i=1}  p(y_{d,i}|f_d(t_{d,i}))  \prod^{D,C}_{d,c=1}\cancel{p(G_{d,c} | \mathbf{v}^G_{d,c})}p(\mathbf{v}^G_{d,c}) \cancel{p(u | \mathbf{v}_u)} p(\mathbf{v}_u)  }{ \prod^{D,C}_{d,c=1}\cancel{p(G_{d,c} | \mathbf{v}^G_{d,c})}q(\mathbf{v}^G_{d,c})\cancel{p(u | \mathbf{v}_u)} q(\mathbf{v}_u)} dS \\
        &= \int \prod^{D,C}_{d,c=1} p(G_{d,c} | \mathbf{v}^G_{d,c})q(\mathbf{v}^G_{d,c})p(u | \mathbf{v}^u)q(\mathbf{v}^u) \\& \hspace{1cm}\times \log \frac{\prod^{D, N_d}_{d,i=1}  p(y_{d,i}|f_d(t_{d,i}))   \prod^{D,C}_{d,c=1}p(\mathbf{v}^G_{d,c})p(\mathbf{v}_u) }{ \prod^{D,C}_{d,c=1}q(\mathbf{v}^G_{d,c})q(\mathbf{v}_u)} dS \\
        &= \int \prod^{D,C}_{d,c=1} p(G_{d,c} | \mathbf{v}^G_{d,c})q(\mathbf{v}^G_{d,c})p(u | \mathbf{v}^u)q(\mathbf{v}^u) \log \prod^{D, N_d}_{d,i=1}  p(y_{d,i}|f_d(t_{d,i})) dS \\ &\hspace{1cm} - \sum^{C, D}_{c, d=1} \text{KL}[q(\mathbf{v}^G_{d,c})||p(\mathbf{v}^G_{d,c})] -\text{KL}[q(\mathbf{v}_u)||p(\mathbf{v}_u)] \\
        &= \mathbb{E}_{q(\{G_{d,c}, \mathbf{v}^G_{d,c}\}^{C, D}_{c, d=1} , u, \mathbf{v}^u) }\Big[\prod^{D, N_d}_{d,i=1}  p(y_{d,i}|f_d(t_{d,i}))\Big] \\ & \hspace{1cm}- \sum^{C, D}_{c, d=1} \text{KL}[q(\mathbf{v}^G_{d,c})||q(\mathbf{v}^G_{d,c})] -\text{KL}[q(\mathbf{v}_u)||p(\mathbf{v}_u)]\\
        &= \sum^{D, N_d}_{d,i=1}\mathbb{E}_{q}[p(y_{d,i}|f_d(t_{d,i}))] - \sum^{C, D}_{c, d=1} \text{KL}[q(\mathbf{v}^G_{d,c})||p(\mathbf{v}^G_{d,c})] -\text{KL}[q(\mathbf{v}_u)||p(\mathbf{v}_u)]\\
    \end{split}
\end{equation*}
where we have used the additive property of the KL divergence. The KL terms for the inducing point distributions are between multivariate Gaussians, with $p(\mathbf{v}^G_{d,c})=\mathcal{N}(0, \mathbf{K}^{G}_{d,c})$ and $p(\mathbf{v}^u)=\mathcal{N}(0, \mathbf{K}^u)$, where the $\mathbf{K}$'s represent the covariance matrices, and $q(\mathbf{v}^G_{d,c})=\mathcal{N}(\boldsymbol{\mu}^{G}_{d,c}, \boldsymbol{\Sigma}^{G}_{d,c})$ and $q(\mathbf{v}^u)=\mathcal{N}(\boldsymbol{\mu}^{u}, \boldsymbol{\Sigma}^{u})$, where the $\boldsymbol{\mu}$'s and $\boldsymbol{\Sigma}$'s represent variational parameters. Using the general result from \citet{gpmlbookpg203}, we obtain
\begin{equation*}
\text{KL}[q(\mathbf{v}^G_{d,c})||p(\mathbf{v}^G_{d,c})]=\frac{1}{2} \log|\mathbf{K}^{G}_{d,c}(\boldsymbol{\Sigma}^{G}_{d,c})^{-1}|+\frac{1}{2}\text{tr}[\mathbf{K}^{G}_{d,c}(\boldsymbol{\mu}^{G}_{d,c}(\boldsymbol{\mu}^{G}_{d,c})^\top + \boldsymbol{\Sigma}^{G}_{d,c} - \mathbf{K}^{G}_{d,c})],
\end{equation*}
and 
\begin{equation*}
\text{KL}[q(\mathbf{v}^u)||p(\mathbf{v}^u)]=\frac{1}{2} \log|\mathbf{K}^u(\boldsymbol{\Sigma}^u)^{-1}|+\frac{1}{2}\text{tr}[\mathbf{K}^u(\boldsymbol{\mu}^u(\boldsymbol{\mu}^u)^\top + \boldsymbol{\Sigma}^u - \mathbf{K}^u)].
\end{equation*}

\end{document}